\documentclass[lettersize,journal]{IEEEtran}
\usepackage{amsmath,amsfonts,amssymb}
\usepackage[ruled]{algorithm2e}
\usepackage{array}
\usepackage{bbm}
\usepackage[caption=false,font=normalsize,labelfont=sf,textfont=sf]{subfig}
\usepackage{textcomp}
\usepackage{stfloats}
\usepackage{url}
\usepackage{verbatim}
\usepackage{graphicx}
\usepackage{booktabs}
\usepackage{multirow} 
\usepackage{cite}
\usepackage{hyperref} 
\usepackage{color}

\makeatletter

\newcommand{\Rmnum}[1]{\expandafter\@slowromancap\romannumeral #1@}
\makeatother

\begin{document}

\title{
Enhancing Social Decision-Making of Autonomous Vehicles: A Mixed-Strategy Game Approach With Interaction Orientation Identification
}

\author{Jiaqi Liu,~\IEEEmembership{Student Member,~IEEE,} Xiao Qi, Peng Hang,~\IEEEmembership{Member,~IEEE,} and Jian Sun
\thanks{This work was supported in part by the National Natural Science Foundation of China (52302502, 52232015), the Young Elite Scientists Sponsorship Program by CAST (2022QNRC001), the Belt and Road Cooperation Program under the 2023 Shanghai Action Plan for Science,Technology and Imnovation (23210750500), the State Key Laboratory of Intelligent Green Vehicle and Mobility under Project No. KFZ2408, the Shanghai Scientific Innovation Foundation (No.23DZ1203400) and the Fundamental Research Funds for the Central Universities.}
\thanks{Jiaqi Liu, Xiao Qi, Peng Hang, and Jian Sun are with the Department of
Traffic Engineering and Key Laboratory of Road and Traffic Engineering,
Ministry of Education, Tongji University, Shanghai 201804, China. (e-mail: \{liujiaqi13, qixaio, hangpeng,  sunjian\}@tongji.edu.cn)}

\thanks{Corresponding author: Peng Hang}
}

\markboth{}%
{Shell \MakeLowercase{\textit{et al.}}: A Sample Article Using IEEEtran.cls for IEEE Journals}

\maketitle 

\begin{abstract}
The integration of Autonomous Vehicles (AVs) into existing human-driven traffic systems poses considerable challenges, especially within environments where human and machine interactions are frequent and complex, such as at unsignalized intersections.
To deal with these challenges, we introduce a novel framework predicated on dynamic and socially-aware decision-making game theory to augment the social decision-making prowess of AVs in mixed driving environments. 
This comprehensive framework is delineated into three primary modules: Interaction Orientation Identification, Mixed-Strategy Game Modeling, and Expert Mode Learning.
We introduce 'Interaction Orientation' as a metric to evaluate the social decision-making tendencies of various agents, incorporating both environmental factors and trajectory characteristics. The mixed-strategy game model developed as part of this framework considers the evolution of future traffic scenarios and includes a utility function that balances safety, operational efficiency, and the unpredictability of environmental conditions. To adapt to real-world driving complexities, our framework utilizes a dynamic optimization framework for assimilating and learning from expert human driving strategies. These strategies are compiled into a comprehensive strategy library, serving as a reference for future decision-making processes.
The proposed approach is validated through extensive driving datasets and human-in-loop driving experiments, and the results demonstrate marked enhancements in decision timing and precision.
\end{abstract}


\section{Introduction}
Autonomous driving technology holds the promise of significantly enhancing the safety and efficiency of the entire transportation system, thereby accelerating its modernization~\cite{wang2023new,li2023safe}. In recent years, the ongoing practical application of autonomous driving technologies has made the integration of autonomous vehicles (AVs) into existing human traffic flows an important and critical issue~\cite{hang2022conflict}. In mixed human-machine driving environments, AVs should not only focus on their own safety requirements but also possess the capability to interact effectively with human-driven vehicles (HVs). Specifically, AVs should learn to understand the driving intentions of other participants and make decisions that are comprehensible to humans ~\cite{zha2023survey, wang2023interacting, lu2022autonomous}. In strongly interactive traffic scenarios ,such as unsignalized intersections, these capabilities are crucial for enhancing driving safety and traffic efficiency, posing higher demands on the decision-making abilities of AVs.

Current autonomous driving decision-making algorithms include rule-based~\cite{zhang2017finite}, deep learning-based~\cite{liu2023mtd,aksjonov2023safety}, and game theory-based methods~\cite{hang2022conflict,cai2021game}. Significant research has been conducted on enhancing the social intent understanding and response capabilities of autonomous driving using these methods~\cite{wang2021socially,peng2021learning,hang2020integrated}.

Compared to rule-based and learning-based decision modeling methods, game theory-based decision methods enable a comprehensive consideration of the interdependencies of decisions made by interacting parties, facilitating dynamic, interactive decision modeling. This approach holds significant potential for improving the social interaction capabilities of autonomous vehicles. 
However, existing game theory-based models in autonomous driving grapple with several challenges: (1) A lack of comprehensive social consideration, where current models primarily focus on trajectory features without fully integrating quantifiable representations of the interaction environment~\cite{ren2019shall,HuWen2023Lane}; (2) The oversight of the ongoing influence of present decisions on future interaction dynamics~\cite{rahmati2017towards,ali2021clacd}, a factor often implicitly considered by human drivers; (3) The complexity in parameter selection for game models, which struggle to adapt to the diverse range of interaction partners and scenarios due to reliance on manually set parameters.

To address these challenges, we propose a dynamic, social decision-making game framework for strong interaction scenarios, comprising three modules: Interaction Orientation Identification, Mixed-Strategy Game Modeling, and Expert Mode Learning.
In the interaction orientation identification module, we define interaction orientation as a concept, quantifying the social decision-making tendencies of interaction subjects by considering the coupled results of environmental and trajectory characteristics. We then develop a mixed-strategy game model that considers the evolution of future states and design a utility function that accounts for safety, efficiency, and random environmental disturbances. Subsequently, we design a dynamic optimization framework for real-world expert strategy learning. The model learns expert interaction patterns from datasets based on this framework, with the learned results stored in an expert strategy library. In the model's decision-making inference process, we dynamically invoke expert strategy parameters based on the identified social tendencies and efficiently solve decisions using the mixed-strategy game model.

Implemented and tested in the challenging context of unsignalized intersections, our framework is subjected to extensive data analysis and human-in-the-loop simulations. It demonstrates superior performance in decision timing, accuracy, and social adaptability compared to various baseline methods.

The contributions of this study are threefold:
\begin{itemize}
    \item Development of a dynamic decision-making framework that augments the social interaction capabilities of AVs. This is achieved by recognizing drivers' social tendencies and constructing a mixed-strategy game model optimized through expert interaction pattern learning.
    \item Introduction of the 'Interaction Orientation' metric, a novel approach for efficiently and dynamically discerning drivers' social styles by amalgamating environmental and trajectory data.
    \item Creation of a game decision model and a dynamic optimization framework for learning from real-world expert strategies. This facilitates efficient and flexible social reasoning and decision-making, with the framework's efficacy validated through extensive empirical experiments and human-in-the-loop simulations.
\end{itemize}

The paper is organized as follows: Section~\ref{section:2} reviews related works. Section~\ref{section:3} describes the decision-making problem and our framework. Section~\ref{section:4} details the proposed decision framework. In Section~\ref{section:5}, data-based experiments and their results are analyzed. And the human-in-loop experiments are analyzed in section ~\ref{section:6}.The paper concludes in Section~\ref{section:7}.

\section{Related Works}
\label{section:2}
\subsection{Decision-Making Methods of AVs}
Intelligent decision-making systems are paramount for the safe and efficient functionality of AVs. The realm of decision-making strategies and algorithms for AVs has witnessed extensive exploration, encompassing rule-based methods\cite{zhang2017finite}, game theory-based methods\cite{cai2021game}, and learning-based methods\cite{peng2021end,liu2023cooperative}.

Rule-based approaches, tailored to specific driving interaction tasks, range from car-following models ~\cite{saifuzzaman2014incorporating,gipps1981behavioural} to lane-changing models ~\cite{gipps1986model,kesting2007general}, and gap acceptance models for intersection behaviors ~\cite{raff1950volume}. However, these methods encounter limitations in complex real-world scenarios. Learning-based methods, such as decision trees ~\cite{barbier2017classification}, LSTM networks ~\cite{phillips2017generalizable}, transformer~\cite{liu2023mtd},and reinforcement learning ~\cite{ronecker2019deep}, face challenges in safety assurance and interpretability.

Game theory, with its inherent capability to analyze interaction and decision-dependence among entities, exhibits a superior understanding of complex interactive decisions and uncertain behaviors. Notable contributions include Kita et al.'s ~\cite{kita1999merging} merging decision model based on mixed-strategy games, Liu et al.'s ~\cite{ban2007game} dual-layer solution framework using genetic algorithms and Nash equilibria, and Hang et al.'s ~\cite{hang2020human} Stackerberg equilibrium model for social behavior integration in merging and overtaking decisions. Additionally, Rahmati et al. ~\cite{rahmati2017towards} developed a dynamic Stackerberg game framework for modeling decisions in unprotected left turns.


\subsection{Driver Characteristics Modeling}
Quantifying and modeling the characteristics of interactive agents is vital for AVs to better recognize these agents and dynamically adjust their decisions based on these characteristics, thereby enhancing decision-making capabilities. Individual behavioral characteristics can be divided into two types: long-term style characteristics and short-term features. The former refers to long-term driving characteristics dominated by personal traits and behavioral styles, while the latter pertains to short-term interactive features influenced by the interactive environment and behaviors of others during driving scenarios ~\cite{chen2020investigating}. Long-term driver characteristics categorize driving style into several types, such as aggressive, normal, and conservative ~\cite{zhu2019typical,hang2020human}, facilitating real-time identification for personalized decision-making ~\cite{zhu2019typical,constantinescu2010driving}. Short-term interactive features are represented by certain characteristic variables, indicating the real-time traits of surrounding interactive agents. For instance, Gaussian mixture regression and continuous hidden Markov models are used for real-time estimation of reference acceleration in car-following states ~\cite{ZhaoJian2022Intelligent}, impact intensity related to acceleration as a driving feature estimation index, and kinematic indicators related to acceleration and speed ~\cite{sagberg2015review}. 

\subsection{Sociality in Decision-Making Behavior Modeling}
For AVs to integrate seamlessly into human-dominated traffic, understanding HV intentions and making human-like decisions is essential. Recent studies have begun integrating social interaction concepts from human sociology into AV-HV interactions\cite{schwarting2019social,wang2021socially,liu10315232}. This includes the introduction of Social Value Orientation (SVO) into game-theoretic utility functions by Schwarting et al.\cite{schwarting2019social}, and Hang et al.'s\cite{hang2022decision} game-theoretic decision framework for unsignalized intersections. Moreover, Wang et al.\cite{wang2021socially} and Behrad et al.\cite{toghi2022social} have developed algorithms for behavioral prediction and MARL frameworks incorporating SVO. Zhao et al.\cite{zhao2021yield} and Peng et al.~\cite{peng2021learning} introduced game-theoretic models emphasizing social preferences and counterfactual reasoning.

Addressing the limitations of solely relying on game-theoretic models, we propose an Interactive Tendency Social Module to capture the dynamic changes in decision-making tendencies of interactive agents. This module aims to enhance the interactive decision-making capabilities of AVs within the framework of game theory.

\section{Problem Statement and Framework Overview}
\label{section:3}
\subsection{Scenario Description}
This research develops and evaluates a decision-making framework, specifically within the context of an unsignalized intersection. Unsignalized intersections, characterized by the absence of traffic signal controls, present a significant challenge for vehicular interaction due to the resultant ambiguous right-of-way scenarios. This complexity necessitates enhanced interactive capabilities for AVs, requiring them to accurately interpret the behavior of other vehicles to make safe and informed decisions.

\begin{figure}
    \centering
    \includegraphics[width=0.8\linewidth]{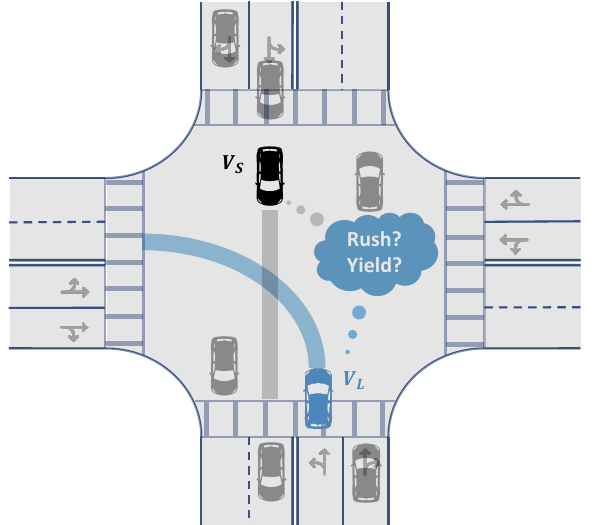}
    \caption{The interaction scenario. }
    \label{fig:scenario description}
\end{figure}

Our study, as depicted in Fig.\ref{fig:scenario description}, focuses on a typical two-way, four-lane cross intersection. We investigate the interaction dynamics with a HV proceeding straight through the intersection. In our model, the driving style, interactive tendencies, and initial conditions of the HV are randomly initialized. The AV, tasked with executing an unprotected left turn, must dynamically gauge the HV's interactive tendencies and adjust its decision-making process in real-time to navigate the intersection safely and efficiently.

\subsection{Framework Overview}


Our proposed social decision-making game framework primarily consists of three modules: driver social preference identification, mixed-strategy game modeling, and expert mode learning. In the driver social preference identification module, we quantify the social decision-making tendencies of interaction objects through interaction tendencies. Subsequently, we construct a mixed-strategy game module that considers future state evolution and design a benefit function that accounts for safety, efficiency, and environmental random disturbances, aiming at achieving safe and efficient interaction decisions.

To enhance the generalization ability and adaptability of social decision-making, we design an expert mode learning module. This module aims to learn expert modes based on real-world data and dynamically adjusts interaction modes during the decision-making process. We develop an expert strategy database where the framework constructs an optimized model to learn human expert strategies and saves the learning outcomes to the expert strategy library. During the application process of the decision model, we utilize the social tendency module to identify the tendencies of interaction objects and search for and load the most stylistically similar expert mode parameters from the expert strategy library, enabling efficient solving of the mixed-strategy social game model.

\begin{figure*}
    \centering
    \includegraphics[width=0.9\linewidth]{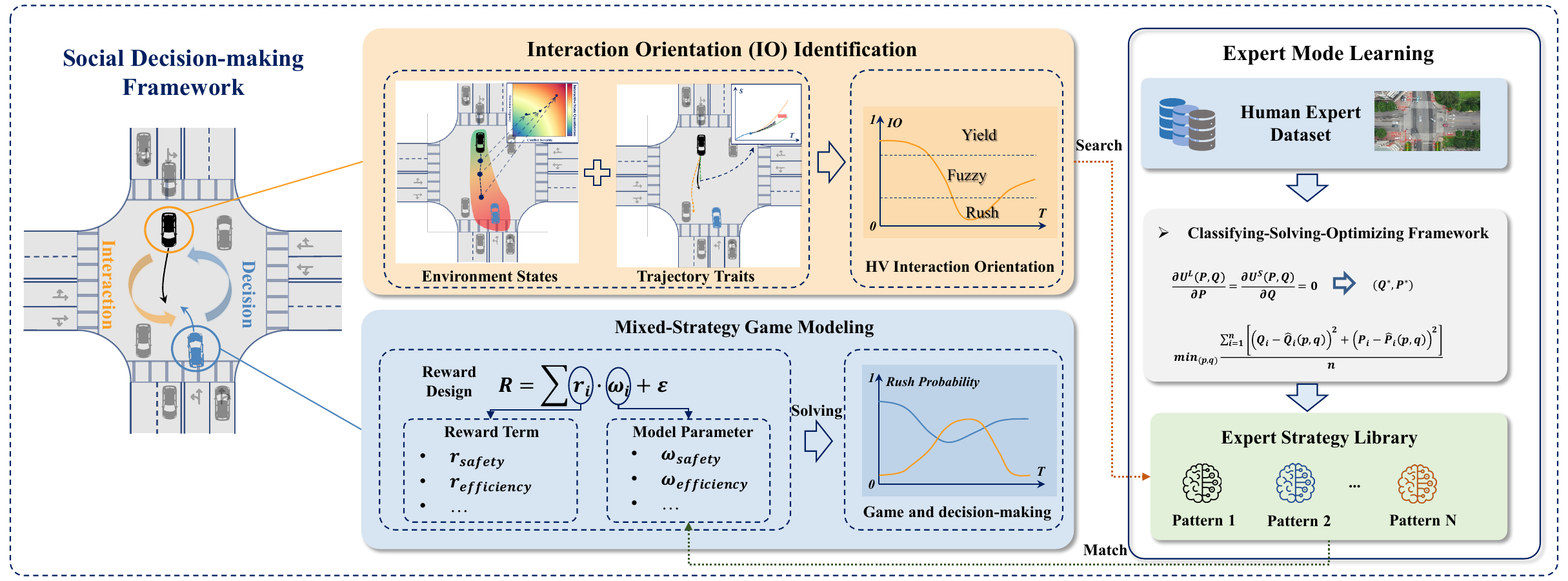}
    \caption{The social decision-making game framework.}
    \label{fig:famework_overview}
\end{figure*}

\section{Methodology}
\label{section:4}

\subsection{Interaction Orientation Identification}
In scenarios where interaction dynamics are complex, the capability of AVs to dynamically discern the social intentions and tendencies of HVs becomes pivotal for making informed, socially coherent decisions. In conceptualizing the recognition of HVs' social tendencies and interaction intentions, two critical factors are considered: first, the interaction intention of an HV can evolve continuously, necessitating dynamic and real-time monitoring; second, HV behavioral patterns are influenced not only by inherent traits but also by environmental contexts, such as the interaction scene. To address these factors, we introduce the concept of Interaction Orientation (IO). IO is defined as a quantified prediction of the probability that an interaction object will exhibit certain behaviors, based on the characteristics observed over a time period, from the observing subject's viewpoint. By integrating the quantification of interaction environment states and trajectory motion characteristics, IO adeptly and dynamically estimates the social tendency of the interaction object.

\subsubsection{Modeling for Interaction Environment State }
In the process of quantifying individual characteristics of interaction objects, integrating environmental states is essential for a more precise quantification of behavior characteristics in specific contexts. We utilize two indicators to formulate the representation of the instantaneous interaction environment state: cooperative acceleration and the time difference to reach the conflict point. The methodologies for calculating these indicators are described as follows:
\begin{equation}
\Delta TTCP = TTCP_{\text{left}} - TTCP_{\text{Straight}}
\end{equation}
where $\Delta TTCP$ represents the time difference for the left-turning vehicle and the oncoming straight vehicle to pass through the conflict point, $TTCP_{\text{left}}$ is the time required for the left-turning vehicle, and $TTCP_{\text{Straight}}$ for the straight vehicle.
\begin{equation}
a_c = \left\{ 
  \begin{array}{ll}
    \frac{2(d_l - v_l T_s)}{T_s^2} & \text{if } d_l \geq 0.5v_l T_s \\
    \frac{v_l^2}{2d_l} & \text{if } d_l < 0.5v_l T_s 
  \end{array} 
\right.
\end{equation}
where $d_l$ and $v_l$ represent the distance and instantaneous speed of the left-turning vehicle to the conflict point, and $T_s$, $d_s$, and $v_s$ are the time required, distance, and speed for the straight vehicle.

For ensuring dimensional consistency and reasonable assessment, we normalize the values of $\Delta TTCP$ and $a_c$ using the $\text{Sigmoid}$ function:

\begin{equation}
TTCP_{\text{norm}} = 1 - \text{Sigmoid}(\alpha_{\Delta TTCP} (\Delta TTCP - \beta_{\Delta TTCP}))
\end{equation}

\begin{equation}
a_{c_{\text{norm}}} = 1 - \text{Sigmoid}(\alpha_{a_c} (\Delta TTCP - \beta_{a_c}))
\end{equation}
\begin{equation}
\text{Sigmoid}(x) = \frac{1}{1 + e^{-x}}
\end{equation}
where $\alpha_i$ is the shape factor that determines the deformation rate, and $\beta_i$ is the design parameter, which indicates the value of the specific index when the normalized result is 0.5. 
Taking the relationship between $\Delta TTCP$ and $\Delta TTCP_{norm}$ as an example, the calculation formulas of $\alpha_{\Delta TTCP}$ and $\beta_{\Delta TTCP}$ are introduced:

\begin{equation}
\left\{
  \begin{array}{ll}
    \alpha_{\Delta TTCP} = \frac{c_1 - c_2}{\Delta TTCP_1 - \Delta TTCP_2} \\[1em]
    \beta_{\Delta TTCP} = \Delta TTCP_1 - c_1 \\[1em]
    c_j = \ln \left( \frac{1 - \Delta TTCP_{\text{norm}}^j}{\Delta TTCP_{\text{norm}}^j} \right), \quad j = 1, 2
  \end{array}
\right.
\end{equation}



Subsequently, we further use the Softmax function to assign weights to $\Delta TTCP_{norm}$ and $a_{c_{norm}}$, expressing them as a Interaction Transient State Index (ITSI):

\begin{align}
\text{softmax}(\Delta TTCP_{\text{norm}}, a_{c_{\text{norm}}}) = \left[ \right. & \frac{e^{\Delta TTCP_{\text{norm}}}}{e^{\Delta TTCP_{\text{norm}}} + e^{a_{c_{\text{norm}}}}}, \\
& \left. \frac{e^{a_{c_{\text{norm}}}}}{e^{\Delta TTCP_{\text{norm}}} + e^{a_{c_{\text{norm}}}}} \right]
\end{align}

\begin{equation}
ITSI = \text{softmax}(\Delta TTCP_{\text{norm}}, a_{c_{\text{norm}}}) \left[ \begin{array}{c} \Delta TTCP_{\text{norm}} \\ a_{c_{\text{norm}}} \end{array} \right]
\end{equation}

\subsubsection{Modeling for Trajectories Motion Feature}

The modeling of trajectory motion during interactive processes is pivotal in reflecting the behavioral tendencies and decision-making intentions of participants. This process involves considering variables such as velocity, acceleration, and deviation. To quantify the trajectory interaction features, we focus on the interaction spatiotemporal occupancy relationship within the Space-Time (ST) trajectory framework.

In this approach, the real-time state of the ST trajectory is calculated, taking into account interaction dynamics. An ST trajectory is essentially a longitudinal displacement path marked with time stamps, depicting the progressive movement characteristics of the vehicle. In our model, the ST trajectory is plotted with the straight vehicle as the primary subject and the left-turning Autonomous Vehicle (AV) as the interaction object. The area occupied within this trajectory framework represents the spatiotemporal relationship between the straight vehicle and the trajectory it follows. This method provides a detailed visualization and analysis of how the straight vehicle occupies space and time within its trajectory, offering insights into its interactive behavior.

\begin{figure*}
    \centering
    \includegraphics[width=0.8\linewidth]{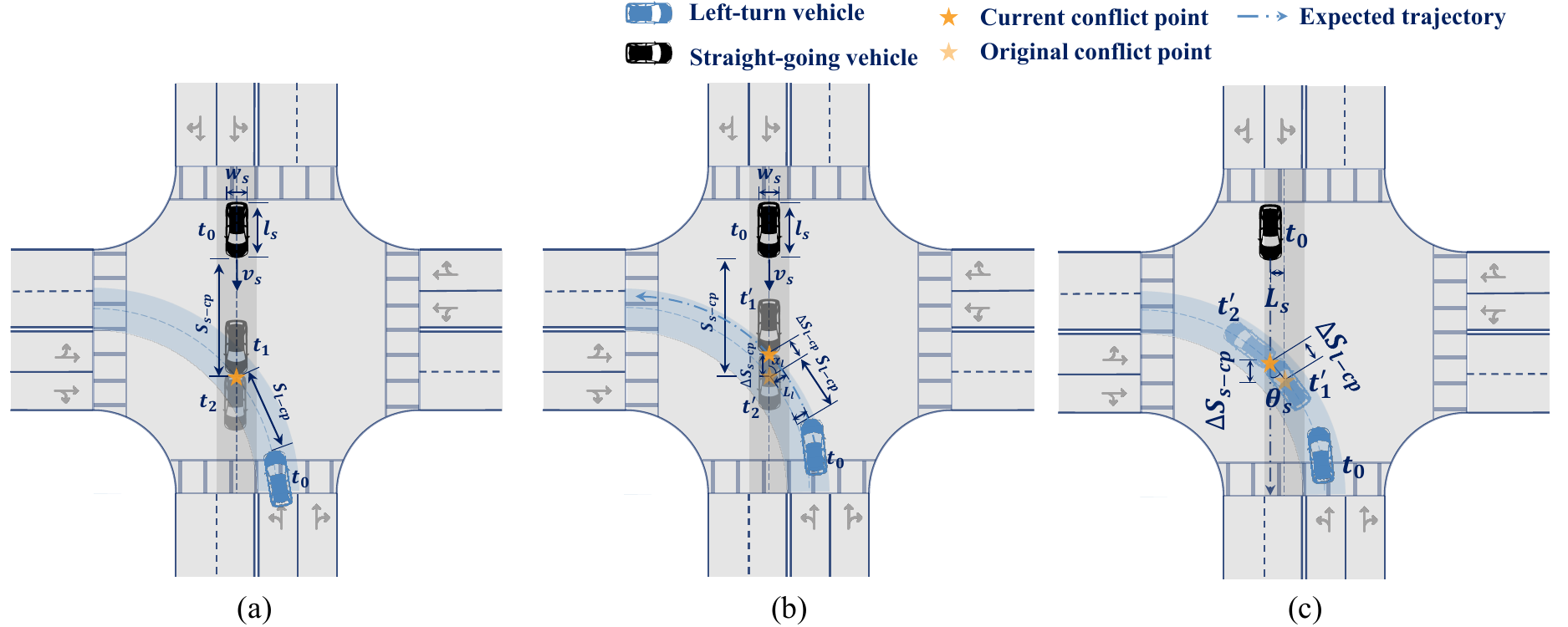}
    \caption{Schematic Diagram of Trajectories for Interacting Parties: (a) Normal Driving, (b) Lateral Deviation in Left-Turning Vehicle, (c) Lateral Deviation in Straight-Going Vehicle.}
    \label{fig:IO_Trajectory_Feature}
\end{figure*}

\begin{figure*}
    \centering
    \includegraphics[width=0.9\linewidth]{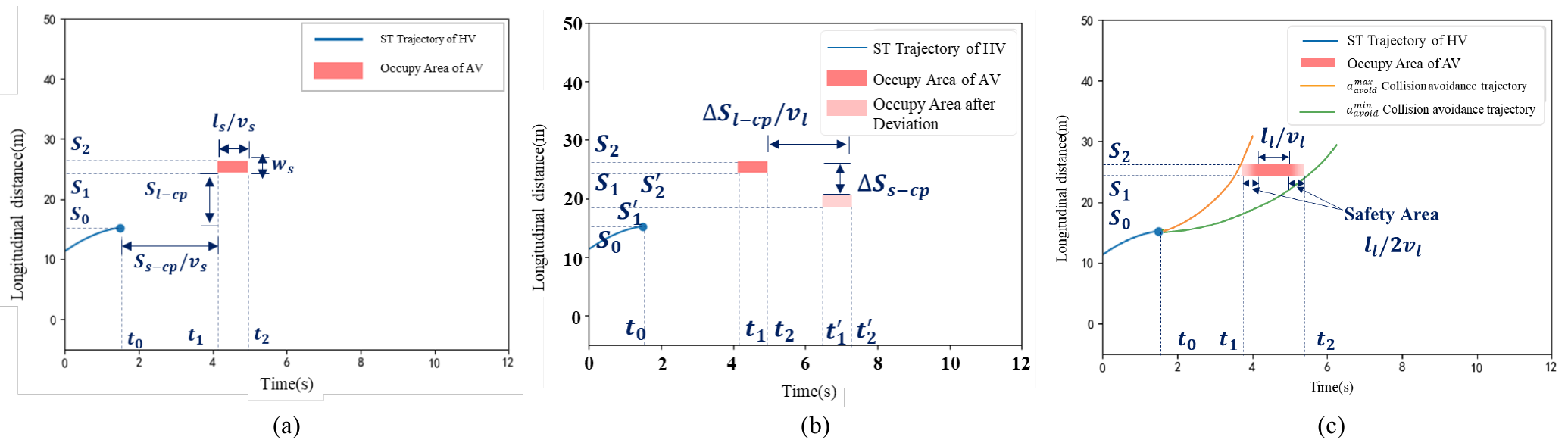}
    \caption{Space-Time (ST) Diagram Centered on the Straight-Going Vehicle: (a) ST Trajectories and Interaction Partner's Occupied Area under Normal Conditions, (b) ST Diagram following Lateral Deviation of the Left-Turning Vehicle, (c) Determining Acceleration Boundaries based on the ST Diagram.}
    \label{fig:IO_ST_Model}
\end{figure*}

As illustrated in Fig.\ref{fig:IO_ST_Model}(a), we have depicted the ST trajectory graph with the straight-going vehicle as the subject. The occupied area represents the temporal and spatial relationship of the straight-going vehicle appearing on its ST trajectory. The red area in the graph indicates the occupied area by the left-turning vehicle, which is determined by the size and instantaneous speed of the left-turning vehicle.
As shown in Fig.4(a), the ST diagram is pictured with the main view of straight-going vehicle. 
The red area denotes the occupied area of left-turn vehicle, which is determined by the size and constant speed of the left-turn vehicle.
If both the straight vehicle and the left-turning vehicle travel along the centerline of their virtual lanes, with the bottom-left coordinate of the occupied area as $t_1, S_1$ and the top-right coordinate as $t_2, S_2$, the coordinates are determined as follows:
\begin{equation}
    \begin{aligned}
     t_1 = t_0 + \frac{S_{l-cp}}{v_l}, \\
    t_2 = \frac{l_l}{v_l} + t_1, \\
    S_1 = S_0 + S_{s-cp},\\
    S_2 = S_1 + w_l + l_l
    \end{aligned}
\end{equation}
where $(t_0,S_0)$ are the current coordinates of the straight vehicle in the ST diagram, $l_l$ and $w_l$ are the length and width of the left-turning vehicle, $v_l$ is its instantaneous velocity, and $S_{l-cp}$ and $S_{s-cp}$ are the longitudinal distances of the left-turning and straight vehicles to the conflict point, respectively, which is shown in Fig.~\ref{fig:IO_Trajectory_Feature}(a).

During interaction, both parties may exhibit lateral deviations. For the left-turning vehicle, a rightward lateral deviation $L_l$ alters the conflict point on the trajectories, increasing the longitudinal distance of the left-turning vehicle to the conflict point by $\Delta S_{l-cp}$ and decreasing the corresponding distance for the straight vehicle by $\delta S_{s-cp}$: 
\begin{equation} 
    \begin{aligned}
    \Delta S_{l-cp} = \tan(\theta_l) \times L_l,  \\
    \Delta S_{s-cp} = \arcsin(\theta_l) \times \Delta S_{l-cp} 
    \end{aligned}
\end{equation} 
where $\theta_l$ is the complement of the angle formed by the tangent at the conflict point on the expected trajectory and the centerline of the straight vehicle lane.

The lateral deviation causes changes in the longitudinal distance of the trajectory intersection point, altering the occupied area on the straight vehicle's ST diagram, as shown in Fig.~\ref{fig:IO_ST_Model}(b). The coordinates of the shifted occupied area are $(t_1',S_1')$ and $(t_2',S_2')$, and their differences from the original coordinates are represented as: 
\begin{equation} 
    \begin{aligned}
    (t_1', S_1') = (t_1 - \Delta t, S_1 + \Delta S),\\
    (t_2', S_2') = (t_2 - \Delta t, S_2 + \Delta S), \\
    \Delta t = \frac{\Delta S_{s-cp}}{v_s}, \\
    \Delta S = \Delta S_{l-cp} 
    \end{aligned}
\end{equation}
Similarly, as illustrated in Fig.~\ref{fig:IO_Trajectory_Feature}(c), if the straight vehicle deviates laterally to the right, the longitudinal distances to the conflict point change accordingly: 
\begin{equation} 
\begin{aligned}
    \Delta S_{l-cp} = \arcsin(\theta_s) \times L_s, \\
    \Delta S_{s-cp} = \arctan(\theta_s) \times L_s 
\end{aligned}
\end{equation} 
where $\theta_s$ is the angle formed by the line connecting the current and original conflict points and the centerline of the straight vehicle lane.

After calculating the real-time state of the ST diagram, we determine the boundary accelerations $[a_{avoid}^{min},a_{avoid}^{max}]$ for the straight vehicle under different decision-making scenarios for collision avoidance. The maximum acceleration for avoidance is determined such that a uniformly accelerating trajectory just passes through the top-left coordinate $(t_1,S_2)$ of the occupied area, while the minimum acceleration is determined to just pass through the bottom-right coordinate $(t_2,S_1)$ of the occupied area, as shown in Fig.~\ref{fig:IO_ST_Model}(c). The calculations are as follows: 
\begin{equation}
\begin{aligned}
    a_{\text{avoid}}^{\text{min}} = \frac{2[(S_1 - S_0) - v_0 (t_2 - t_0)]}{(t_2 - t_0)^2}, \\ 
    a_{\text{avoid}}^{\text{max}} = \frac{2[(S_2 - S_0) - v_0 (t_1 - t_0)]}{(t_1 - t_0)^2} 
\end{aligned}
\end{equation}
After calculating the boundary accelerations, we normalize the motion behavior trajectory features. Assuming the current velocity of the straight vehicle at time $t_0$ is $v_s$, the minimum and maximum longitudinal distances traveled under the boundary accelerations $[a_{avoid}^{min},a_{avoid}^{max}]$ within the time domain $[t_0,t_s]$ are as follows: 
\begin{equation} 
\begin{aligned}
    S_{\text{min}} = -\frac{((v_s + a_{\text{avoid}}^{\text{min}} (t_s - t_0))^2 - v_s^2)}{2a_{\text{avoid}}^{\text{min}}}, \\
S_{\text{max}} = \frac{((v_s + a_{\text{avoid}}^{\text{max}} (t_s - t_0))^2 - v_s^2)}{2a_{\text{avoid}}^{\text{max}}} 
\end{aligned}
\end{equation}


Based on the minimum longitudinal distance $S_{max}$ and maximum longitudinal distance $S_{min}$ obtained above, compared to the actual longitudinal displacement $\Delta S$ produced by the real trajectory within the time domain $[t_0,t_s]$, we calculate the normalized value of the motion behavior trajectory feature $S_{norm}$ as follows: 
\begin{equation}
S_{\text{norm}} = \left\{
  \begin{array}{ll}
    1 & \text{if } (\Delta S \geq S_{\text{max}}) \\
    \frac{\Delta S - S_{\text{min}}}{S_{\text{max}} - S_{\text{min}}} & \text{if } (S_{\text{min}} < \Delta S < S_{\text{max}}) \\
    0 & \text{if } (\Delta S \leq S_{\text{min}})
  \end{array}
\right.
\end{equation}

where the minimum longitudinal distance \(S_{max}\) and the maximum longitudinal distance \(S_{min}\) are determined based on Equation (15), with \(\Delta S_{long}\) being the actual longitudinal displacement distance produced by the real trajectory within the time domain \([t_0, t_s]\).

\subsubsection{Interaction Orientation Calculation}
In determining the IO, we integrate two pivotal factors: the ITSI and the normalized trajectory feature $S_{norm}$. The formula for calculating IO is structured as follows:
\begin{equation} 
IO = 1 - ITSI \times S_{norm} 
\label{eq:IO_cal}
\end{equation} 

\begin{figure}
    \centering
    \includegraphics[width=0.8\linewidth]{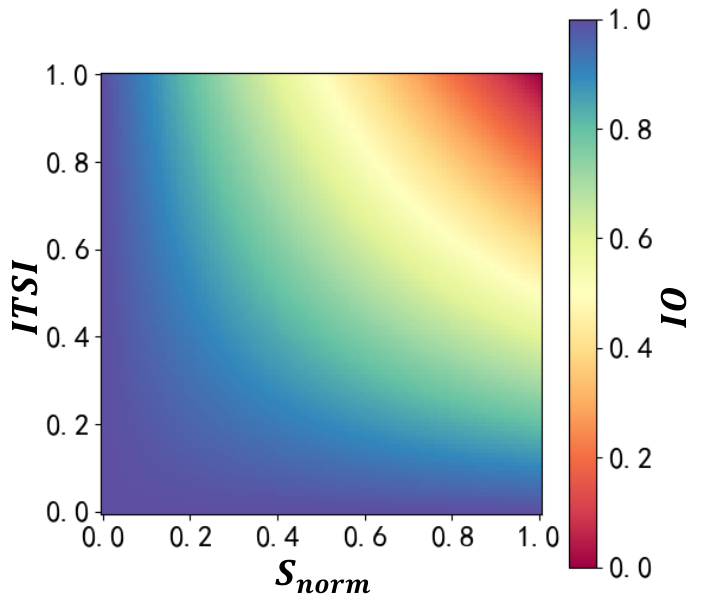}
    \caption{The relationship between IO and ITSI and $S_{norm}$}
    \label{fig:IO Description}
\end{figure}

According to this equation, both ITSI and $S_{norm}$ undergo a normalization process, ensuring that the resulting IO value ranges between $[0,1]$. In this context, a value of 0 denotes a pronounced precedence tendency, whereas a value of 1 signifies a distinct yielding tendency, as illustrated in Fig.\ref{fig:IO Description}. It is observed that when ITSI remains constant, an increase in $S_{norm}$ corresponds to a decrease in IO, signifying a more pronounced precedence tendency. Conversely, a reduction in $S_{norm}$ under a constant ITSI increases IO, indicating a stronger yielding tendency. Similarly, for a constant $S_{norm}$, an increase in ITSI results in a reduced IO, reflecting a stronger precedence tendency, and vice versa.

\subsection{Mixed-Strategy Game Modeling}
In certain complex interactive scenarios, such as unprotected left turns, there is no absolute right of way for the interacting participants, and the sequential order of dynamic games is difficult to determine. Considering the dynamic and uncertainty of the interaction process, as well as the potential reasoning of human drivers about future risks and situation evolution, we model the interaction between participants as a non-cooperative mixed-strategy game, incorporating future behavior deductions, and design a payoff function that considers safety, efficiency, and uncertain disturbances.

\subsubsection{Game Problem Modeling}
Game Problem Modeling Suppose the game participants are the left-turning vehicle $V_l$ and the straight-going vehicle $V_s$, represented by the set $I={V_l,V_s}$. The strategy sets are $S={S_l,S_s}$, with the strategy set for $V_l$ being $S_l={s_l^p,s_l^y}$, and the interaction decision for Vs represented as $S_s={s_s^p,s_s^y}$, where $s_l^p$ is the strategy to proceed, and sly is to yield. The payoffs for both sides are denoted as $u_l(S_l,S_s)$ and $u_s(S_l,S_s)$.

In the mixed-strategy game, the strategy set of each participant is a sample space, denoted as $\Delta (S_i)$ for $S_i$ (where i=l,s):
\begin{equation}
\Delta(S_i) = \{ p_i = (p_i^p, p_i^y), p_i^j \geq 0, \sum_j p_i^j = 1 \}
\end{equation}
where $i$ represents the interacting participant and $j$ the decision strategy. The mixed strategy is denoted as $p_i=(p_i^p,p_i^y)\in \Delta (S_i)$, and the outcome of the mixed-strategy game between both parties is $p=(p_l,p_s)$, with $p_i\in \Delta(S_i)$.

The expected payoff for the participants is:
\begin{equation}
U_i(p) = \sum_{s \in S} p(s) u_i(s) = \sum_{s = (s_l, s_s) \in S} p_l(s_l) \times p_s(s_s) \times u_i(s_l, s_s)
\end{equation}

If the mixed-strategy game outcome $p=(p_l,p_s)$ is a mixed-strategy Nash equilibrium, for the expected payoffs $U_l$ and $U_s$ of the left-turning and straight-going vehicles, respectively, it always holds that:
\begin{equation}
\left\{
  \begin{array}{ll}
    U_l(p_l, p_s) \geq U_l(p_l', p_s) \\
    U_s(p_l, p_s) \geq U_s(p_l, p_s')
  \end{array}
\right.
\end{equation}
where $p_l^{\prime} \in \Delta(S_l)$ and $p_s^{\prime}\in \Delta(S_s)$, and each participant chooses the optimal strategy probability distribution given the unchanging strategy of the other participant.
Specifically, when $V_l$ chooses a strategy $s_l$ from $S_l$ with probability distribution $p_l$, and $V_s$ chooses $s_s$ from $S_s$ with $p_s$, the expected payoffs for $V_l$ and $V_s$ using mixed strategies are:
\begin{equation}
\left\{
  \begin{array}{ll}
    u_l(s_l^p, p_s) = \sum_j p_s^j u_l(s_l^p, s_s^j) \\
    u_l(s_l^y, p_s) = \sum_j p_s^j u_l(s_l^y, s_s^j) \\
    u_s(p_l, s_s^p) = \sum_j p_l^j u_s(s_l^j, s_s^p) \\
    u_s(p_l, s_s^y) = \sum_j p_l^j u_s(s_l^j, s_s^y)
  \end{array}
\right.
\end{equation}
where j is in the strategy space and $j\in (p,y)$. The total expected payoffs for $V_l$ and Vs under the mixed strategy combination $p=(p_l,p_s)$ are respectively:
\begin{equation}
\left\{
  \begin{array}{ll}
    U_l(p) = p_l^p u_l(s_l^p, p_s) + p_l^y u_l(s_l^y, p_s) \\
    U_s(p) = p_s^p u_s(p_l, s_s^p) + p_s^y u_s(p_l, s_s^y)
  \end{array}
\right.
\label{eq:payoff_matrix}
\end{equation}

\subsubsection{Payoff Function Design}
The design of the payoff function is crucial for the performance of game-theoretic decision models. Human interactions not only consider immediate benefits like safety and efficiency but also involve reasoning about future states. Therefore, we first model the future state extrapolation of environmental risks and incorporate efficiency, safety, and stochastic disturbance benefits into the payoff function $u_i(S)$: 
\begin{equation}
u_i(S) = \alpha_i^S T_i^S + \beta_i^S R_i^S + \epsilon_i^S
\label{eq:payoff_cal}
\end{equation}
Here, $\alpha_{iS}$ denotes the coefficient for efficiency benefits $T_{iS}$, $\beta_{iS}$ for safety benefits $R_{iS}$, and $\epsilon_{iS}$ represents normally distributed stochastic disturbance benefits. Both $\alpha_{iS}$ and $\beta_{iS}$ are parameters to be solved based on real human interactive decision-making.

(a) \quad Future State Extrapolation.
Assume that the parties accelerate at $a_i^p$ initially and decelerate at $a_i^y$ until they reach a conflict point or a velocity threshold. Let the time interval between the current moment and the future state extrapolation be $\Delta t$, the future state decision $j$ can be expressed as:
\begin{equation}
\left\{
  \begin{array}{ll}
    v' = v_0 + a_i^j \Delta t \\
    d' = d_0 - \frac{(v' + v_0)}{2} \Delta t
  \end{array}
\right.
\end{equation}
where \( v_0, d_0 \) are the current velocity and distance to the conflict point, respectively, and \( v', d' \) are the future extrapolated velocity and distance. \( j \) can be either \( p \) or \( y \). 

In the process of future state motion extrapolation, if the velocity reaches a threshold, the vehicle maintains a constant velocity. The threshold values for the velocity states are: 
\begin{equation}
    \begin{cases} 
    v_l^{\min} \leq v_l' \leq v_l^{\max} \\ v_s^{\min} \leq v_s' \leq v_s^{\max} 
    \end{cases}
\end{equation}
where \( v_l', v_s' \) are the extrapolated velocities for left-turning and straight-going vehicles, respectively. \( v_l^{\min}, v_l^{\max}, v_s^{\min}, v_s^{\max} \) are the actual minimum and maximum velocities during the turning and straight-going processes at intersections. Multiple-step continuous extrapolation is represented as \( n\Delta t \) (\( n \in \mathbb{N}^+ \)), where \( N \) is the number of extrapolation steps.

(b) \quad Safety Benefits.
Safety benefits are characterized by the time difference to reach the conflict point, $\Delta 
 TTCP$, and the cooperative acceleration $a_c$. When the game decision is set as $S={S_l,S_s}$, the $\Delta TTCP_i^{(S,n \Delta t)}$ and $a_{c,i}^{S,n \Delta t}$ for the interacting participants at a future moment $n \Delta t (n\in N^+)$ can be represented as: 
 \begin{equation}
\Delta TTCP_i^{S,n\Delta t} = TTCP_i^{S,n\Delta t} - TTCP_{-i}^{S,n\Delta t}
\end{equation}
\begin{equation}
TTCP_i^{S,n\Delta t} = \frac{d_i^{S,n\Delta t}}{v_i^{S,n\Delta t}}
\end{equation}
\begin{equation}
a_{c,i}^{S,n\Delta t} = \frac{2(d_i^{S,n\Delta t} - v_i^{S,n\Delta t} TTCP_{-i}^{S,n\Delta t})}{(TTCP_{-i}^{S,n\Delta t})^2}
\end{equation}
where subscript $-i$ indicates the interaction partner of participant $i$, $TTCP_i^{(S,n \Delta t)}$ is the time to reach the conflict point, $d_i^{S,n \Delta t}$ and $v_i^{S,n \Delta t}$ are the distance and velocity, respectively. From this, $\Delta TTCP_i^{(S,n \Delta t)}$ $a_{c,i}^{S,n \Delta t}$ for both parties can be obtained and normalized to $\Delta TTCP_{i_{norm}}^{S,n \Delta t}$ and $a_{c,i_{norm}}^{S,n \Delta t}$. Let $R_i^{S,n \Delta t}$ represent the safety benefit for participant $i$ at that moment, it can be expressed as: 
\begin{equation}
R_i^{S,n\Delta t} = \text{softmax}(\Delta TTCP_{i_{\text{norm}}}^{S,n\Delta t}, a_{c,i_{\text{norm}}}^{S,n\Delta t}) \left[ \begin{array}{c} \Delta TTCP_{i_{\text{norm}}}^{S,n\Delta t} \\ a_{c,i_{\text{norm}}}^{S,n\Delta t} \end{array} \right]
\end{equation}

The total safety benefit $R_i^S$ considering both current and future safety benefits can be calculated as: 
\begin{equation}
R_i^S = \sum_n \gamma^n R_i^{S,n\Delta t} \quad (n=1,2,3,\ldots)
\end{equation} 
where $n$ denotes the number of future state time steps considered, and $\gamma \in (0,1)$ is the future state discount factor.

(c) \quad Efficiency Benefits.
The study considers time delays under different decisions as efficiency benefits. The efficiency benefit $T_i^S$ for participant $i$ under game decision $S={S_l,S_s}$ is designed as: 
\begin{equation}
T_i^S = \frac{2}{1 + e^{(TTCP_{i,0} - TTCP_i^S)}} - 1
\end{equation}


Here, \(TTCP_{(i,0)}\) is the time required for participant \(i\) to pass the conflict point under the current state, and \(TTCP_i^S\) is the time required under decision \(S\). The calculation of \(TTCP_i^S\) needs to consider the impact of different decisions on behavior: if participant \(i\) chooses to preempt, they move to the conflict point with uniform acceleration \(a_i^{preempt}\); if yielding, they decelerate uniformly to a stop with \(a_i^{yield}\), wait for the interaction partner \(-i\) to pass, and then proceed with \(a_i^{preempt}\). 
Thus, \(TTCP_i^S\) is represented as:

\begin{equation}
TTCP_i^S = \left\{
  \begin{array}{ll}
    \frac{-v_0 + \sqrt{v_0^2 + 2a_i^p d_0}}{2}, & \text{if } S_i = s_i^p \\
    \frac{\sqrt{2a_i^p (d_0 - \frac{v_0^2}{2a_i^y})}}{2} + TTCP_{-i,0}, & \text{if } S_i = s_i^y
  \end{array}
\right.
\end{equation}

(d) \quad Stochastic Disturbance Benefits.
To minimize the impact of random disturbances on decision determination, assume a standard normal distribution $X\sim N(0,1)$. The stochastic disturbance benefit $\epsilon_i^S$ in this study is distributed as:
\begin{equation}
\epsilon_i^S \sim N(\mu, \sigma^2), \quad X = \frac{\epsilon_i^S - \mu}{\sigma} \sim N(0, 1)
\end{equation}

\subsection{Expert Mode Learning}
The performance of game decision models is critically influenced by the adept learning and setting of parameters. Traditional approaches often employ manual parameter setting or rely on fixed calibration configurations, which limit the models' adaptability and responsiveness in diverse environments. To overcome these limitations, we have developed the Expert Mode Learning module that derives driving patterns from real expert data.

In this framework, a real-world expert dataset $\mathcal{D}$ is preprocessed, facilitating the extraction and categorization of various expert driving patterns. Interaction events are segregated into three categories based on their interaction orientation: precedence tendency, ambiguous tendency, and yielding tendency, represented as $d_1$, $d_2$, and $d_3$ respectively. For each category, we initiate an iterative learning process. This process starts with parameter and model initialization, followed by constructing the participant payoff matrix, and advancing to optimization solving. Each iteration produces a set of expert parameters ${\alpha_i^S,\beta_i^S }$. The payoff function of interaction participants $u_i (S)$ is then calculated. Utilizing the Lemke-Howson algorithm\cite{shapley2009note}, linear complementarity problems are adjusted and solved to approximate mixed-strategy Nash equilibriums, determining the current predicted decisions of left-turning vehicles $\hat{p}_l^n$ and straight vehicles $\hat{p}s^n$, compared with the real labels $P_l^n$ and $P_s^n$. The learning objective focuses on minimizing the discrepancy between algorithmic solutions and actual observed decisions, as defined by the following objective function:
\begin{equation}
\min \left( \frac{\sum{n=1}^{N} [(P_l^n - \hat{p}_l^n)^2 + (P_s^n - \hat{p}_s^n)^2]}{N} \right)
\label{eq:opti_target}
\end{equation}
Here, $n$ represents the interaction segment number within a certain pattern category, and $N$ denotes the total number of segments in this category.

The iterative learning process is optimized using a heuristic algorithm, specifically the Genetic Algorithm (GA)~\cite{mirjalili2019genetic}. Upon convergence of the iterations, the expert strategy parameters for that pattern are finalized, concluding the algorithm. If convergence is not achieved, the process iterates further. The end product is the expert parameters $e^i={\alpha_i^{S*},\beta_i^{S*} }$ corresponding to different strategies $s^i$, stored in the expert strategy repository $E={e^{s^i},s^i \in S}$.
This learning process is outlined in Algorithm \ref{alg:expert_learning}.

During the inference application stage, at any given time step t, the relevant expert pattern is identified and matched based on the interaction orientation identification result. After loading the expert parameters, the decision model is resolved using the Lemke-Howson algorithm~\cite{shapley2009note}, implementing the decision outcome. This procedure is repeated until the interaction sequence concludes.

\begin{algorithm}
\SetAlFnt{\small}
\SetKwInOut{Parameter}{Inputs}
\SetKwInOut{Output}{Output}
\caption{Expert Mode Parametric Learning}
\label{alg:expert_learning}
\LinesNumbered
\SetAlgoLined
\Parameter{Expert Dataset $\mathcal{D}$}
\Output{Expert Strategy Library $E$}
\hrule
\vspace{0.2em}

Pre-classify the expert data set $\mathcal{D}$ based on interaction tendencies to obtain $\mathcal{D} = \{d_1, d_2, d_3\}$;\\
Initialize the expert strategy library $E$;\\
\For{each $d_i$ in $\mathcal{D}$}{
    Initialize decision parameter set $Pa_i^0$ and game payoff matrix $M_i^0$;\\
    \For{each $j$ in $N_i$}{
        \For{each $t_j$ in $T_j$}{
            Get environmental state information $ITSI$ and trajectory motion characteristics $S_{norm}$;\\
            Calculate $IO_i$ based on Eq.\ref{eq:IO_cal};\\
            Calculate the current payoff matrix $M_i^{t_j}$ based on Eq.\ref{eq:payoff_cal} and Eq.\ref{eq:payoff_matrix};\\
            Calculate optimal strategy probability distribution $p_i^{t_j}$ using the Lemke-Howson algorithm;\\
            Compute loss error $Loss_i^{(t_j)}$ based on Eq. \ref{eq:opti_target};\\
            \If{$Loss_i^{t_j} < \delta$}{
                Obtain $Pa_i^*$, Break;\\
            }
            \Else{
                Use GA for parameter optimization to obtain the updated $Pa_i^{(t_j)}$;\\
            }
        Update expert strategy library $E = E \cup Pa_i^*$;\\
        }
    }
}
\end{algorithm}

\section{Data-based Experiments and Analysis}
\label{section:5}
This section introduces the datasets utilized in our experiment, proceeds with the validation of the interaction orientation identification model, and concludes with an analysis and discussion of the solution results of the game decision model.

\subsection{Dataset and Parameter Setting}
\textbf{Dataset.} For model learning and validation, the Argoverse2 Motion Dataset~\cite{wilson2023argoverse} and SinD datasets~\cite{xu2022drone} are employed. The Argoverse2 Motion Dataset, amassed by Argo AI across six cities in the United States, encompasses approximately 763 hours and includes over 440,000 scene groups rich in interactive scenarios. The SinD dataset, gathered by Tsinghua University at an unsignalized intersection in Tianjin, China, documents the trajectory data of more than 4,800 vehicles. From the Argoverse2 dataset, we select strong interaction scenarios involving left-turn and straight-on traffic at unsignalized intersections to validate the interaction orientation identification module. In the SinD dataset, 268 interaction events are extracted, which include 132 scenarios where left-turning vehicles take precedence, 136 where they yield, and the number of segments for straight-on traffic yielding, exhibiting no clear tendency, and proceeding first are 421, 800, and 362, respectively. This data serves for expert strategy learning.

\textbf{Parameter setting.}
In the design of the benefit function for the mixed-strategy model, for equations (24) and (25), we set the parameters as follows: \(\Delta t=0.1\), \(v_l^{\min}=0 \, \text{m/s}\), \(v_l^{\max}=5 \, \text{m/s}\), \(v_s^{\min}=0 \, \text{m/s}\), \(v_s^{\max}=10 \, \text{m/s}\). For equation (30), we set \(n=5\) and \(\gamma=0.5\). For equation (33) and the corresponding random disturbance process: following the \(3\sigma\) rule of normal distribution to minimize the impact of random disturbances on decision certainty, we set \(\mu=0\) and \(3\sigma=0.01\), ensuring that \(P\{\left|\epsilon_i^S\right| \leq 0.01\} = 0.9974\).

\begin{figure}
    \centering
    \includegraphics[width=1\linewidth]{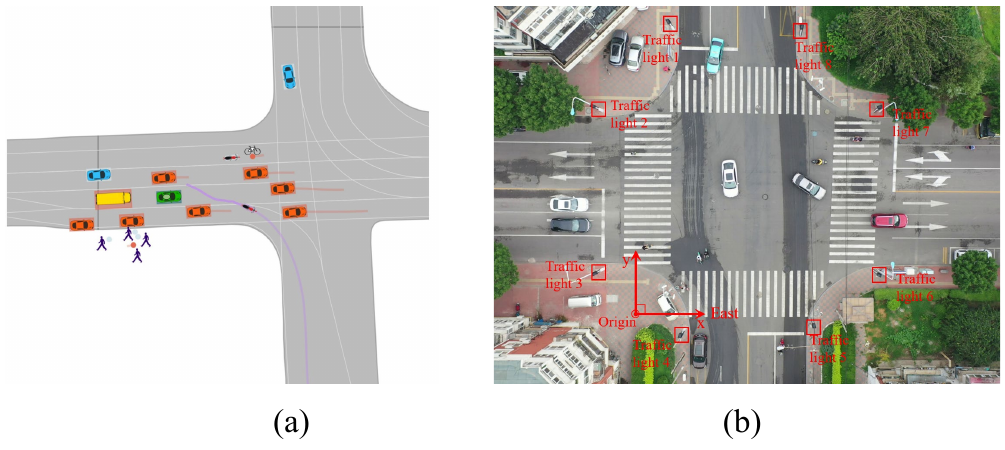}
    \caption{The intersection data from two datasets used to model and testify our model, (a)the Argoverse2 Motion Dataset, (b) the SinD Dataset.}
    \label{fig:dataset}
\end{figure}

\subsection{Interaction Orientation Module Validation}
\begin{figure}
    \centering
    \includegraphics[width=1\linewidth]{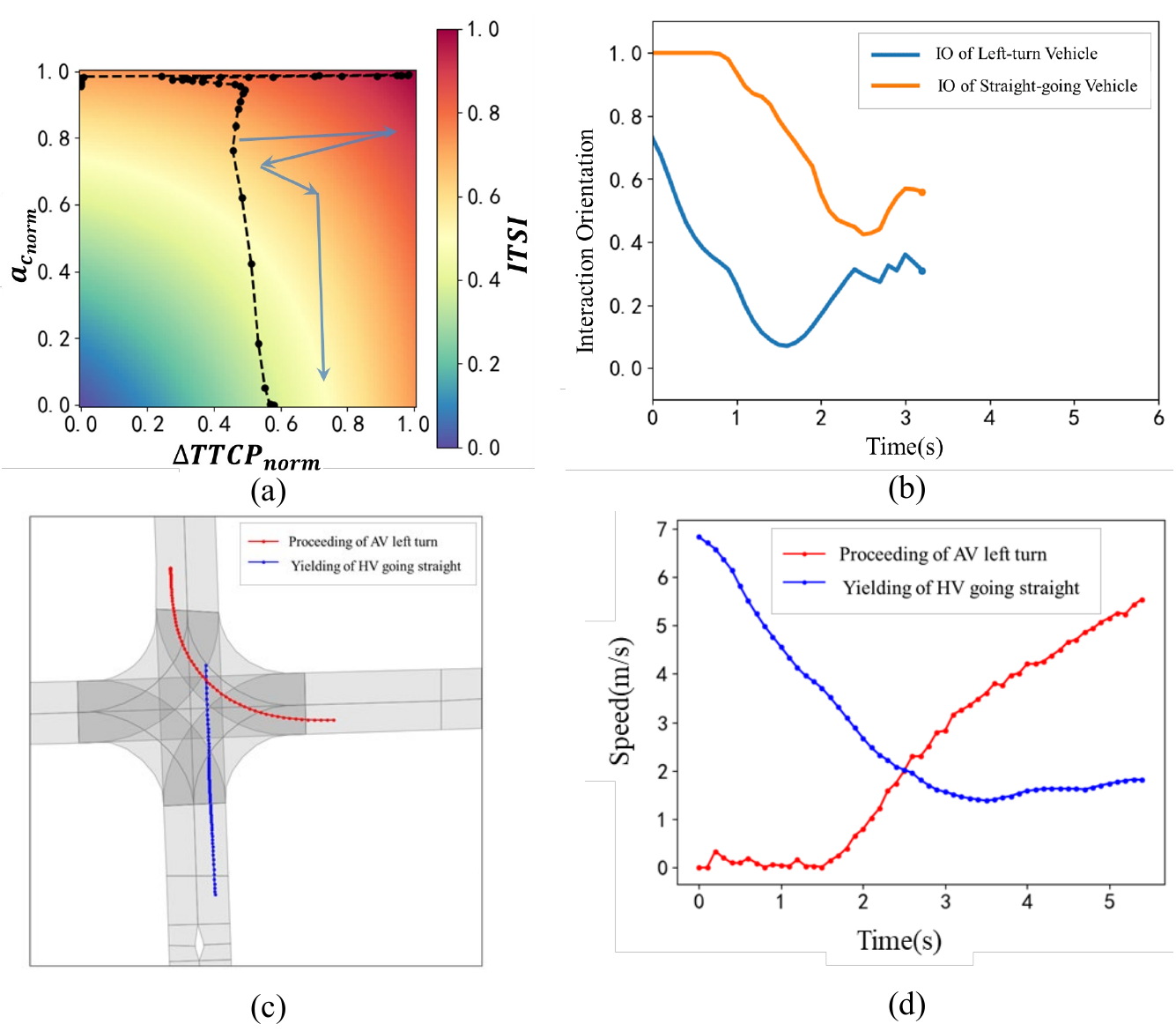}
    \caption{The motion and interaction information of the typical scenario, (a) the quantification result of interaction environment state, (b)the interaction orientation of turn-left vehicle and going-straight vehicle in the scenario,(c) the ground truth trajectories of two vehicles in this scenario, (d) the speed curve of two vheicles during the interaction progress. }
    \label{fig:IO_result}
\end{figure}

A typical unprotected left-turn interaction scenario from the Argoverse Motion dataset is analyzed using our interaction tendency model. The Interaction Transient State Index (ITSI) results for both the left-turning and straight-on vehicles at each moment are depicted in Fig.\ref{fig:IO_result}(a). ITSI trends over time, marked by blue arrows, show an initial increase followed by a gradual decrease. The interaction tendencies of both parties at each moment, as illustrated in Fig.\ref{fig:IO_result}(b), reveal the straight-on vehicle consistently exhibits a yielding tendency, while the left-turning vehicle's tendency progressively shifts from yielding to advancing. Fig.~\ref{fig:IO_result}(c) and (d) display the actual behaviors captured in the dataset, where the left-turning vehicle initially stops to yield and then accelerates to proceed, and the straight-on vehicle transitions from decelerating to yield to slowly approaching the conflict point. This behavior pattern, as calculated by our interaction orientation module, aligns with the recorded data, further validating the module's effectiveness in quantifying interaction tendencies.

\subsection{Expert Mode Learning Results}
\begin{figure}
    \centering
    \includegraphics[width=1\linewidth]{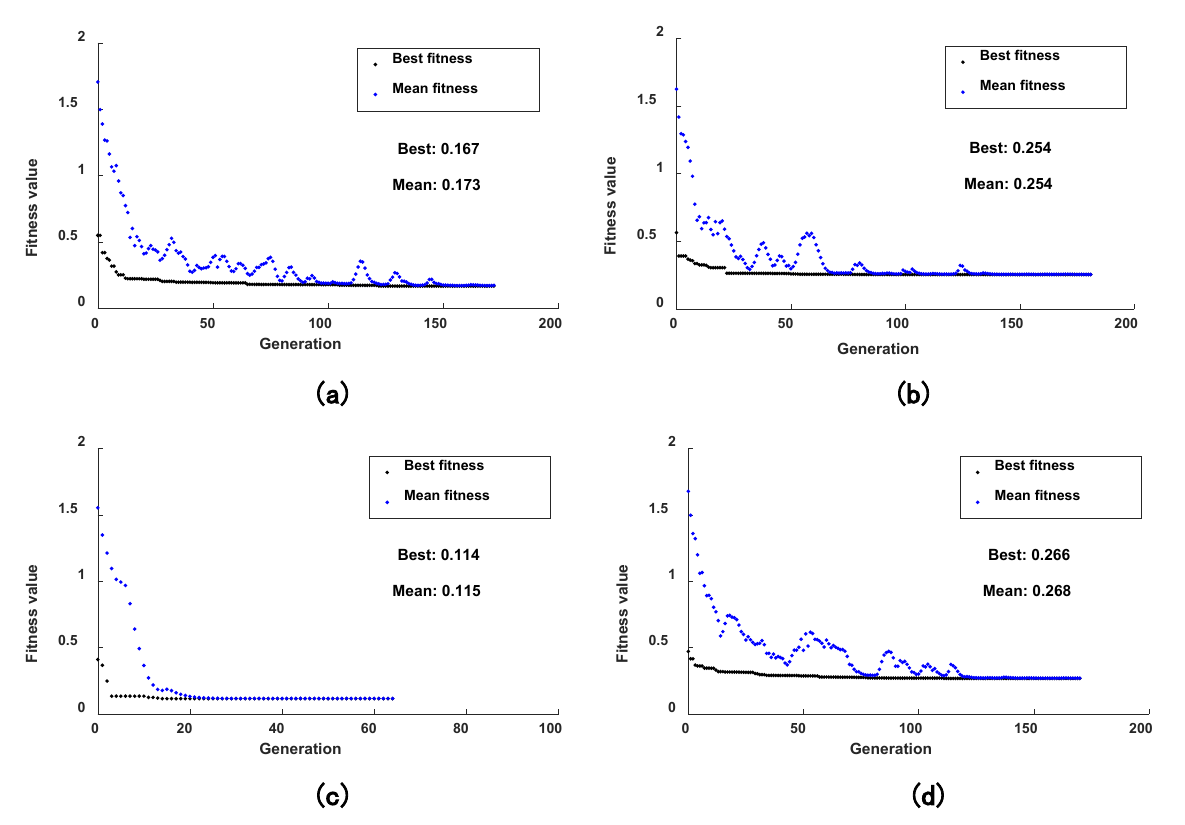}
    \caption{The results comparison of parameters solving between our model and normal game model, (a)our model for solving the proceeding tendency scenarios,(b)our model for solving the ambiguous tendency scenarios,(c)our model for solving the yielding tendency scenarios, (d)normal game model for all scenarios.}
    \label{fig:para_solving}
\end{figure}

This subsection focuses on analyzing the learning outcomes of our expert model based on the original dataset. We juxtapose the model derived from expert data with its counterpart learned without expert data, showcasing their respective learning curves in Fig.~\ref{fig:para_solving}. The final error rates of our expert learning model for scenarios where the straight-on vehicle proceeds first, yields, or shows no clear tendency stand at 0.167, 0.114, and 0.254, respectively. In contrast, the non-expert learned game model registers a final error rate of 0.266.
The findings highlight several key points:
\begin{itemize}
    \item The solution errors of the expert learning model are consistently lower than those of the non-expert learning game model;
    \item Notably, when the oncoming straight-on vehicle exhibits a proceeding or yielding tendency, our model markedly outperforms the non-expert game model. Specifically, the decision model accuracy improves by 37.2\% under a proceeding tendency and by 57.1\% under a yielding tendency. Even in instances of no clear tendency, the decision model accuracy experiences a modest enhancement of 4.5\%, culminating in an overall error reduction of 23\% in our decision model's parameter solution compared to the non-expert game model;
    \item These results underscore the enhanced capability of our model to accurately interpret the intentions of an oncoming straight-on vehicle, whether yielding or proceeding, thereby facilitating more precise decision-making.
\end{itemize}

Beyond solution accuracy, prediction accuracy rate is another metric reflecting the model's precision. A comparison between the prediction outcomes of the expert learning model and the non-expert learning model is presented in Tab.\ref{tab:model_prediction_accuracy}. Our expert learning decision model exhibits a prediction accuracy rate of 83.0\% for segments involving left-turn proceeding and straight-on yielding, and 93.3\% for segments with left-turn yielding and straight-on proceeding, both figures surpassing those of the non-expert learning model.

\begin{table*}[h]
\centering
\caption{Comparison of Prediction Accuracy Between This Study's Decision Model and the Standard Game Theory Model}
\label{tab:model_prediction_accuracy}
\begin{tabular}{@{}lcccccc@{}}
\toprule
\textbf{Decision Scenario} & \textbf{Actual Quantity} & \multicolumn{2}{c}{\textbf{Our Decision Model}} & \multicolumn{2}{c}{\textbf{Standard Game Theory Model}} \\ \cmidrule(lr){3-4} \cmidrule(lr){5-6}
 &  & \textbf{Quantity} & \textbf{Accuracy} & \textbf{Quantity} & \textbf{Accuracy} \\ \midrule
(Left Turn Yields, Straight Goes First) & 749 & 622 & 83.0\% & 603 & 80.5\% \\
(Left Turn Goes First, Straight Yields) & 837 & 781 & 93.3\% & 718 & 85.8\% \\ \bottomrule
\end{tabular}
\end{table*}


\subsection{Decision Model Effectiveness Analysis}
This subsection is devoted to evaluating the practical application efficacy of our decision model. For this purpose, we compare it with five baseline algorithms: Decision Tree Model, Logit Model, Support Vector Machine (SVM), Long Short-Term Memory Networks (LSTM), and Mixed Strategy Game Model.

\subsubsection{Decision Accuracy}
Each of the five aforementioned methods, along with our model, undergoes training and calibration using identical datasets. Their performance is then evaluated and compared, with the results presented in Tab.\ref{tab:decision accuracy}. The LSTM and Logit models demonstrate relatively modest performance, achieving an overall prediction accuracy rate of approximately 70\%. The Decision Tree, SVM, and Mixed Strategy Game models exhibit improved accuracy, with rates hovering around 80\%. Our model outperforms these alternatives, attaining an overall accuracy rate close to 90\%, thereby underscoring the superiority of our approach.

\begin{table*}[h]
\centering
\caption{Comparison of Decision Accuracy in Interaction Scenarios}
\resizebox{\textwidth}{!}{%
\label{tab:decision accuracy}
\begin{tabular}{@{}lccccccc@{}}
\toprule
\textbf{Decision Scenario} & \textbf{Total} & \textbf{Decision Tree} & \textbf{Logit} & \textbf{SVM} & \textbf{LSTM} & \textbf{Mixed Strategy Game Model} & \textbf{Our Method} \\ \midrule
(Left Turn Goes First, Straight Yields) & 749 & 606 (80.1\%) & 555 (74.1\%) & 613 (81.8\%) & 524 (70.0\%) & 598 (79.8\%) & 622 (83.0\%) \\
(Left Turn Yields, Straight Goes First) & 837 & 715 (85.4\%) & 589 (70.4\%) & 704 (84.1\%) & 585 (69.9\%) & 697 (83.2\%) & 781 (93.3\%) \\ \bottomrule
\end{tabular}%
}
\end{table*}

\subsubsection{Decision Timing Analysis}
\begin{figure}
    \centering
    \includegraphics[width=0.8\linewidth]{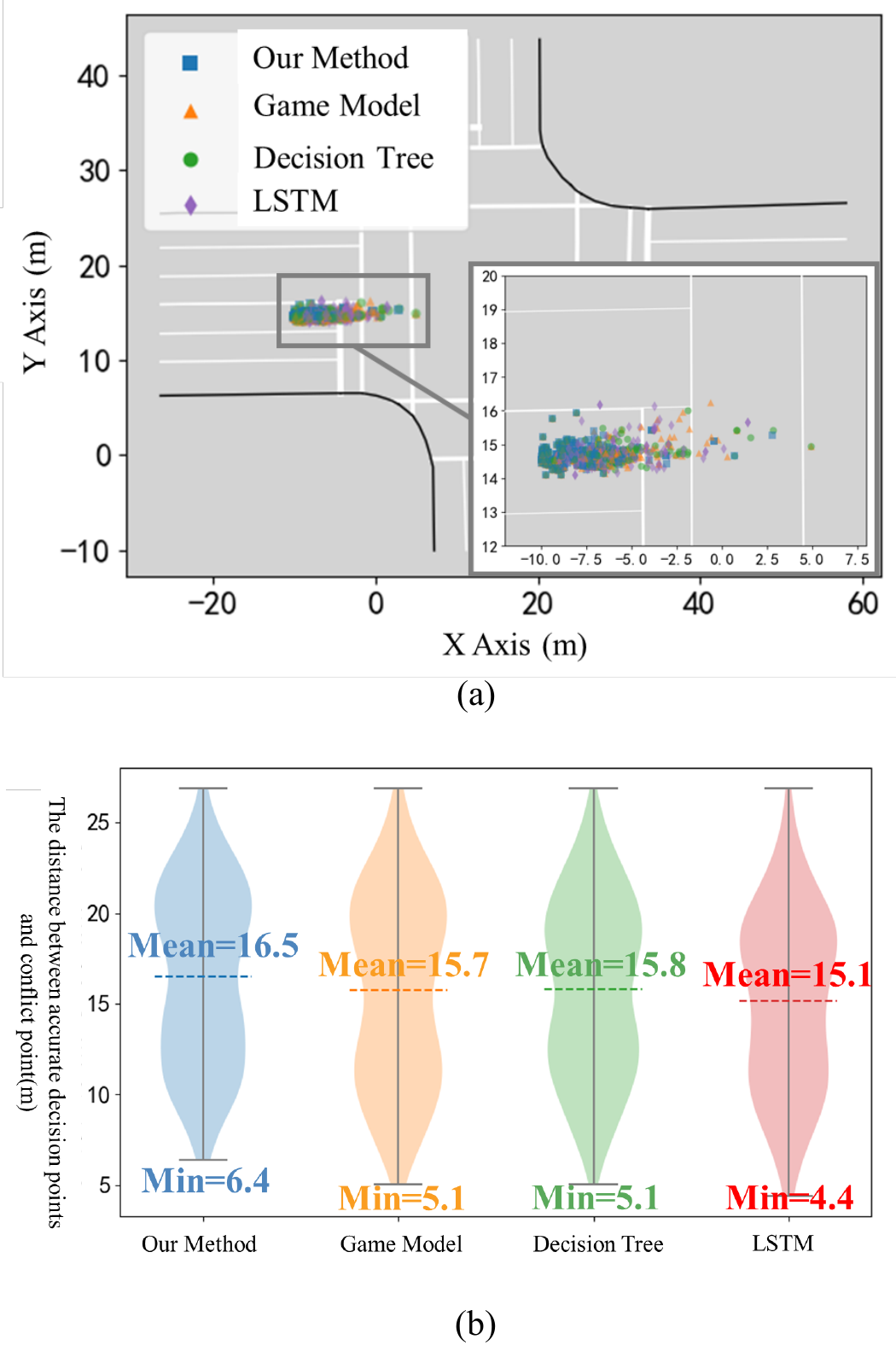}
    \caption{The comparison results of decision timing from different methods, (a)the location distribution of accurate decision points, (b) the distance distribution from accurate decision points to conflict point. }
    \label{fig:Decision_timing_analysis}
\end{figure}

The timeliness of decision-making in interactive processes is pivotal, impacting interaction risk, urgency, and the experiential quality for both parties involved. Delayed decision timing can escalate interaction risk due to ambiguous intentions. We employ the accurate decision point location – the earliest instance where the model’s decision aligns with the actual outcome – as a measure to compare the timeliness of decision-making across different models. This comparison is focused on their proficiency in capturing the precise timing for decisions during interactions. The distribution of accurate decision points and their proximity to the conflict point for various models are illustrated in Fig.~\ref{fig:Decision_timing_analysis} (a) and (b). Our decision method manifests an average distance of 16.5m to the accurate decision point, compared to 15.7m, 15.8m, and 15.1m for the other methods, respectively. This suggests that our method is capable of making more prompt and accurate decisions at a greater distance from the conflict point.

\subsection{Case Study: Interaction-Orientation Model Efficacy}

This subsection presents a case study from the SinD dataset, comparing the performance of our Interaction-orientation model (IO-model) with a Non-IO baseline model in realistic interactive driving scenarios.

\begin{figure*}
\centering
\includegraphics[width=0.8\linewidth]{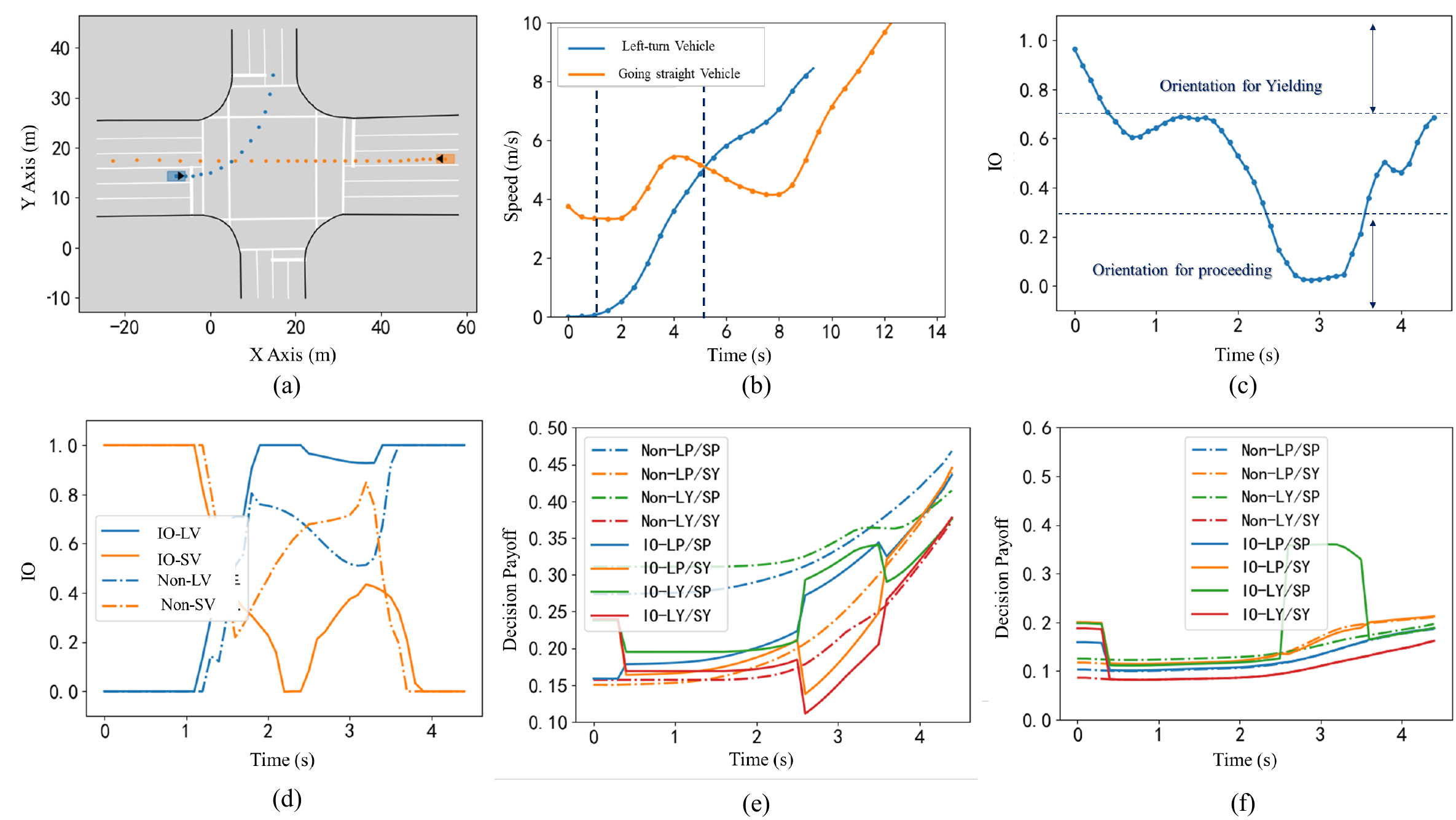}
\caption{Motion states and interaction decision progression in the case study: (a) trajectories of both vehicles, (b) speed curves, (c) interaction orientation of the straight-going vehicle, (d) proceeding probability, (e) game payoff for the left-turn vehicle, (f) game payoff for the straight-going vehicle. LP and LY denote the proceeding and yielding decisions for the left-turn vehicle, while SP and SY indicate the same for the straight-going vehicle.}
\label{fig:case_analysis}
\end{figure*}

Fig. \ref{fig:case_analysis}(a) illustrates the initial state and motion paths of the vehicles involved, while Fig. \ref{fig:case_analysis}(b) depicts their speed variations. In the highlighted scenario, one vehicle approaches the intersection at a reduced speed, reaching the conflict point in 5.5 seconds, whereas the opposing vehicle maintains a speed over 3.5 m/s.

As the left-turn vehicle enters the intersection, it begins real-time analysis of the approaching vehicle's interaction tendencies. These tendencies shift from initial yielding to ambiguity, then to a proceeding inclination, and back to ambiguity, as recorded over time. Our IO-model dynamically adjusts its strategies based on these changing tendencies, utilizing a library of expert patterns.

Decision probabilities illustrated in Fig. \ref{fig:case_analysis}(b) reveal that until 1.1 seconds, both models predict the same outcomes. However, from 1.1 to 2.5 seconds, our IO-model adapts, predicting a 100\% proceeding probability for the left vehicle by 2 seconds. The left vehicle's right-of-way is maintained above 90\% after this point, with the opposing vehicle predominantly yielding. A notable shift occurs between 2.5 and 3.6 seconds, as the straight vehicle accelerates, altering its decision from yielding to 40\% proceeding. The consensus reached is the left vehicle proceeding and the straight vehicle yielding, influenced by their respective acceleration and deceleration patterns.

In contrast, the Non-IO model shows a slower response, with decision probabilities aligning only in the latter stages of interaction (around 3.8 seconds), demonstrating a lag in adapting to dynamic tendencies compared to our IO-model.

The comparison of benefits for both vehicles under various decisions, as computed by the two models (Fig. \ref{fig:case_analysis}(c) and (d)), indicates that the left-turn vehicle's benefits increase under LP/SP strategy when the straight vehicle's tendency changes to proceeding. For the straight vehicle, significant benefits are observed only in the LY/SP decision under the same condition. These shifts in benefits validate our model's effectiveness in adapting decision probabilities based on observed interaction tendencies.

\section{Validation on the Human-in-loop Driving Platform}
\label{section:6}
In this section, a human-in-the-loop driving simulation platform is developed to evaluate the efficacy of a novel decision-making algorithm. The platform was utilized to conduct simulations with human participants, examining the algorithm's performance in terms of efficiency, safety, and decision-making consistency in social contexts.

\subsection{Platform Construction and Experimental Design}
\begin{figure}
    \centering
    \includegraphics[width=1\linewidth]{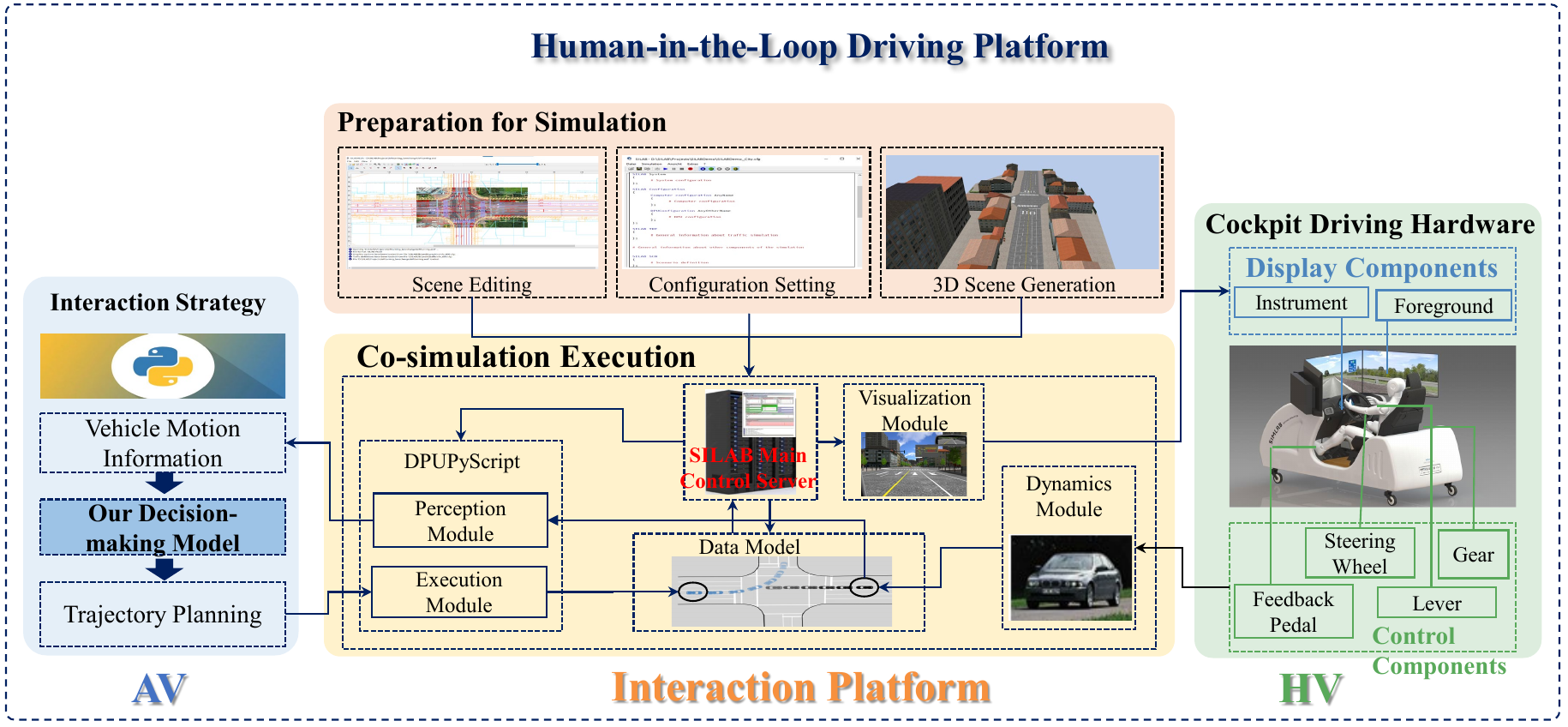}
    \caption{The human-in-loop driving platform.}
    \label{fig:human-in-loop driving platform}
\end{figure}

The experimental platform, depicted in Fig.\ref{fig:human-in-loop driving platform}, integrates a combination of hardware, software, and interactive strategies. A total of 14 participants were recruited, each engaging in 20 unique driving scenarios.
The simulated environment consisted of a four-lane intersection. Vehicles controlled by the algorithm turned left from the west while participant-controlled vehicles drove straight from the east, facilitating interaction. For comparison, a standard non-interactive game-theoretic approach was employed to control left-turn vehicles.

\subsection{Experimental Findings}
The experiment focused on gathering trajectory and speed data of participants and algorithm-controlled vehicles. These data were analyzed for efficiency, safety, and decision-making consistency.
\subsubsection{Efficiency Analysis}
\begin{figure}
    \centering
    \includegraphics[width=0.9\linewidth]{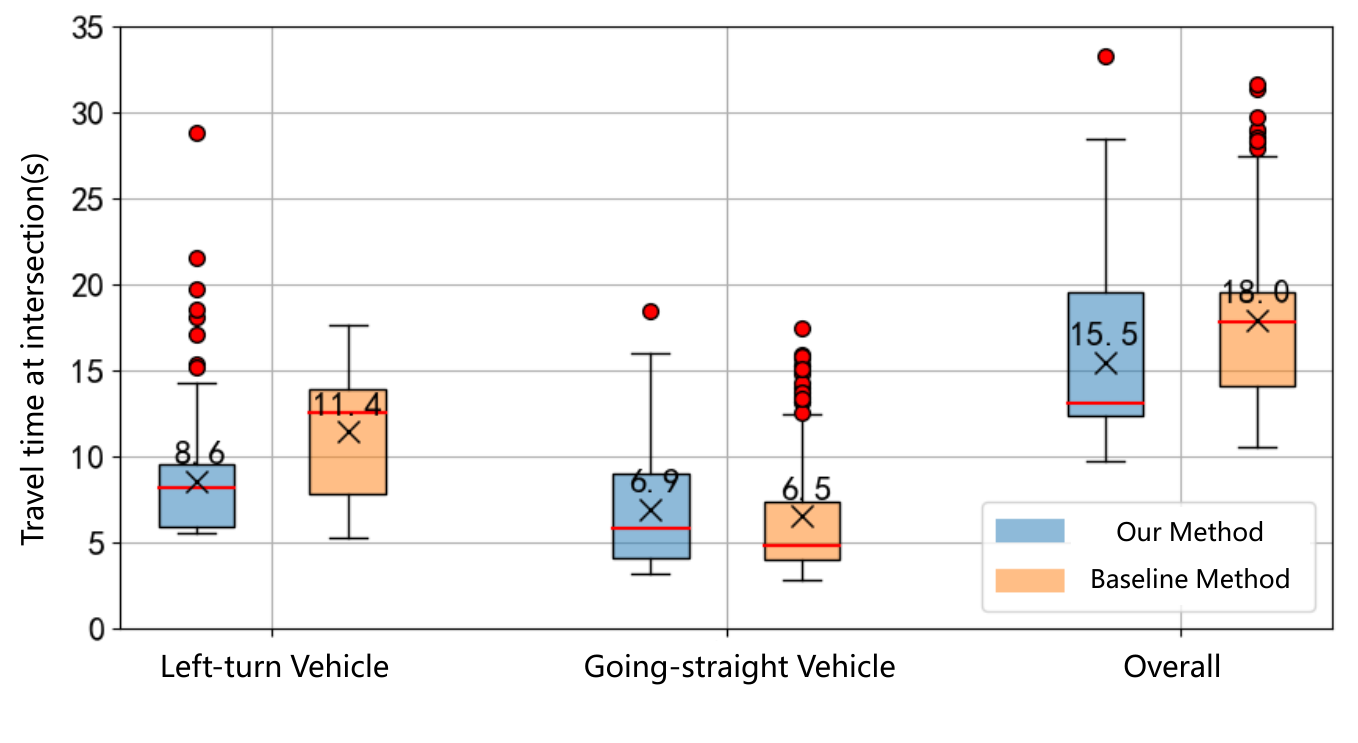}
    \caption{The travel time of two vehicles.}
    \label{fig:human_travel_time}
\end{figure}

Transit times within the intersection were used to gauge efficiency. The interaction strategies impacted both the individual and combined efficiency of the vehicles. Comparative analysis between left-turning AV, straight-going HV, and their combined transit times revealed significant differences. Fig.\ref{fig:human_travel_time} and Tab.\ref{tab:transit_time_stats} detail these findings. Notably, left-turning AV under our algorithm showed a reduced transit time (mean 8.6s) compared to the baseline strategy (mean 11.4s, p=0.000). Straight-going HV showed a marginal increase in transit time with our method (mean 6.9s) versus the control (mean 6.5s, p=0.553). Overall, the combined transit time under our method (mean 15.5s) was significantly lower than the control (mean 18.0s, p=0.000), indicating an overall enhancement in traffic efficiency.

\begin{table}[h]
\centering
\caption{Descriptive Statistics of Transit Times Under Different Strategies}
\label{tab:transit_time_stats}
\resizebox{\columnwidth}{!}{%
\begin{tabular}{@{}lcccc@{}}
\toprule
\textbf{ } & \textbf{Median(s)} & \textbf{Mean (s)} & \textbf{Test Statistic W} & \textbf{p-value} \\ \midrule
Left Turn AV (n=140) & 8.2 (12.5) & 8.6 (11.4) & 1279.0 & 0.000 \\
Straight HV (n=140) & 5.8 (4.8) & 6.9 (6.5) & 4385.0 & 0.553 \\
Overall (n=140) & 13.1 (17.8) & 15.5 (18.0) & 2680.0 & 0.000 \\ \bottomrule
\end{tabular}%
}
\textit{Note: In the Median and Mean columns, the values outside and inside the parentheses represent our method and the comparison method, respectively.}
\end{table}

\subsubsection{Safety Evaluation}
\begin{figure}
    \centering
    \includegraphics[width=0.75\linewidth]{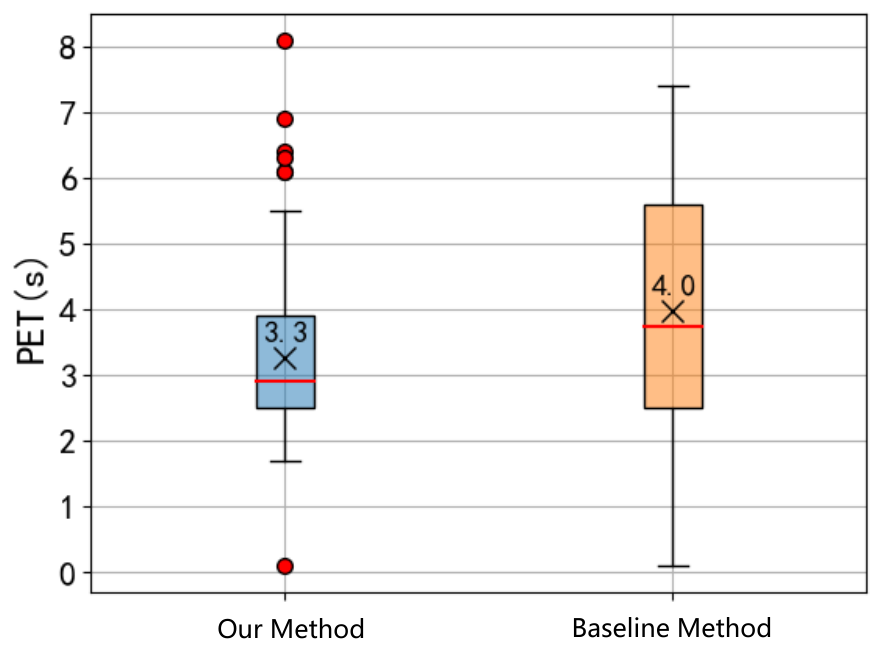}
    \caption{The distribution results of PET.}
    \label{fig:human_in_loop_PET}
\end{figure}

Safety, a critical aspect of vehicle interaction, was assessed using Post-Encroachment Time (PET) as an indicator of conflict severity during left-turn-straight interactions. PET distributions are presented in Fig.\ref{fig:human_in_loop_PET} and Tab.\ref{tab:pet_stats}. Our strategy resulted in a lower mean PET (3.3s) compared to the control (4.0s, p=0.000), suggesting improved safety. Despite a superficial indication of reduced safety in our strategy, a higher frequency of higher PET ($>$5s) events in the control strategy highlighted its inefficiencies and misunderstandings in AV interactions. Further safety assessments, including collision and severe conflict events (PET$<$2s), are detailed in Tab.\ref{tab:collision_conflict_stats}. Our strategy demonstrated superior safety, with lower rates of collisions and severe conflicts compared to the control, underscoring its efficacy in ensuring safe interactions between AV and HV.

\begin{table}[h]
\centering
\caption{PET Statistical Results Under Different Strategies}
\label{tab:pet_stats}
\resizebox{\columnwidth}{!}{%
\begin{tabular}{@{}lcccc@{}}
\toprule
\textbf{Strategy} & \textbf{Median Mdn} & \textbf{Mean Mean} & \textbf{Test Statistic W} & \textbf{p-value} \\ \midrule
Our Strategy (n=140) & 2.9s & 3.3s & 2742.5 & 0.000 \\
Baseline Strategy (n=140) & 3.75s & 4.0s & & \\ \bottomrule
\end{tabular}%
}
\end{table}

\begin{table}[h]
\centering
\caption{Statistics of Collision and Severe Conflict Events}
\label{tab:collision_conflict_stats}
\resizebox{\columnwidth}{!}{%
\begin{tabular}{@{}lcc@{}}
\toprule
\textbf{Strategy} & \textbf{Collision Events} & \textbf{Severe Conflict (PET$<$2s) Events} \\ \midrule
Our Strategy (n=140) & 1 (0.7\%) & 5 (3.6\%) \\
Baseline Strategy (n=140) & 2 (1.4\%) & 13 (9.3\%) \\ \bottomrule
\end{tabular}%
}
\end{table}

\subsubsection{Decision-Making Consistency Assessment}
We also evaluate the decision-making consistency of interaction strategies in a human-in-the-loop driving simulation, focusing on the congruence of decisions between participants and the algorithm across various scenarios.

A total of 28 trials involving 14 participants were conducted, assessing both subjective perceptions and objective outcomes of decision-making. As shown in Tab.\ref{tab:decision_consistency}, the study found that our proposed strategy achieved over 90\% decision-making consistency, a significant improvement of more than 20\% compared to the control strategy. This was evident across most scenarios, except those with ambiguous boundaries where consistency slightly dropped to 78.6\%. This drop was attributed to evenly matched advantages leading to interaction conflicts and impeding clear decision-making.

Further analysis revealed minimal variance between subjective perceptions and objective outcomes under our strategy, demonstrating its accuracy and effective communication of intentions. In contrast, the comparison strategy showed significant discrepancies in these aspects, particularly in scenarios of ambiguous boundaries and left-turn advantages, indicating a deficiency in decision accuracy and intent expression.

\begin{table*}[h]
\centering
\caption{Decision Consistency Analysis Results}
\label{tab:decision_consistency}
\begin{tabular}{@{}lcccccc@{}}
\toprule
\textbf{Interaction Strategy} & \multicolumn{5}{c}{\textbf{Decision Consistency Scenario Proportion (each scenario n=28)}} & \textbf{Total} \\ \cmidrule(lr){2-6}
 & \textbf{Boundary Ambiguous} & \textbf{Straight More Adv.} & \textbf{Straight Adv.} & \textbf{Left Turn More Adv.} & \textbf{Left Turn Adv.} &  \\ \midrule
\multicolumn{7}{@{}l}{\textit{Subjective Feeling (Do you think the left-turn car's decision conflicts with yours?)}} \\
Our Strategy & 78.6\% & 89.3\% & 100\% & 96.4\% & 96.4\% & 92.1\% \\
Baseline Strategy & 46.5\% & 75.0\% & 82.1\% & 67.9\% & 64.3\% & 67.1\% \\
\multicolumn{7}{@{}l}{\textit{Objective Result (AV decision output matches HV decision choice)}} \\
Our Strategy & 82.1\% & 96.4\% & 100\% & 100\% & 96.4\% & 94.9\% \\
Baseline Strategy & 72.0\% & 96.4\% & 96.4\% & 48.1\% & 51.9\% & 73.3\% \\
\bottomrule
\end{tabular}%
\end{table*}



\section{Conclusion}
\label{section:7}
In the mixed human-machine traffic environment, the social interaction and decision-making capabilities of AVs are of great importance. To better understand the decision-making tendencies of interaction partners, a mixed-strategy game decision-making framework is proposed that incorporates the identification of interaction tendencies. This framework is divided into three modules: interaction orientation identification , mixed-strategy game modeling, and expert mode learning. We identify interaction tendencies based on environmental interactions and vehicle trajectory features and further optimize social interaction by learning from real-world expert modes. Using the typical scenario of an unsignalized intersection, the experimental framework is designed. Learning and validation are conducted using various real-world datasets, proving that our approach surpasses existing baseline methods in terms of decision timing, accuracy, and social impact.

In future endeavors, our objective is to broaden the application of our methodology to encompass more intricate and dynamic contexts. This expansion includes addressing the social decision-making processes of AVs when confronted with multiple interacting entities or conglomerations of vehicles. Such an advancement aims to augment the adaptability and generalizability of our algorithm significantly. Furthermore, we intend to undertake the extraction and assimilation of prior knowledge and potential norms governing social interactions from an extensive compilation of datasets derived from real-world experts. This initiative is designed to enrich and enhance the capabilities of autonomous driving algorithms in social decision-making contexts.


\bibliographystyle{IEEEtran} 
\bibliography{reference}

\end{document}